\pdfoutput=1

\documentclass[11pt]{article}

\usepackage[]{ACL2023}

\usepackage{times}
\usepackage{latexsym}
\usepackage{graphicx}
\usepackage{amsmath}
\usepackage{amsfonts}
\usepackage{pdfpages}

\usepackage[T1]{fontenc}

\usepackage[utf8]{inputenc}
\usepackage{CJKutf8}
\usepackage[russian,english]{babel}
\usepackage{microtype}

\usepackage{inconsolata}

\usepackage{float}
\usepackage{subfigure}
\usepackage{multirow}

\definecolor{DSgray}{cmyk}{0,1,0,0}

\title{Leveraging Multi-lingual Positive Instances in Contrastive Learning to Improve Sentence Embedding}

\author{Kaiyan Zhao\thanks{\hspace{1.5mm}Equal contribution.} ,
        Qiyu Wu$^{*}$,
        Xin-Qiang Cai,
        Yoshimasa Tsuruoka \\
       \normalfont{The University of Tokyo}, Tokyo, Japan  \\ 
       \normalfont{\fontsize{11pt}{12pt}\selectfont {\fontfamily{qcr}\selectfont \{zhaokaiyan1006, qiyuw, caixq, yoshimasa-tsuruoka\}@g.ecc.u-tokyo.ac.jp}}
}

\newcommand{\MN}{MPCL}
\newcommand{\blue}[1]{\textcolor{black}{#1}}

\begin{document}
\maketitle

\begin{abstract}
Learning multi-lingual sentence embeddings is a fundamental task in natural language processing. Recent trends in learning both mono-lingual and multi-lingual sentence embeddings are mainly based on contrastive learning (CL) among an anchor, one positive, and multiple negative instances. In this work, we argue that leveraging multiple positives should be considered for multi-lingual sentence embeddings because (1) positives in a diverse set of languages can benefit cross-lingual learning, and (2) transitive similarity across multiple positives can provide reliable structural information for learning.
In order to investigate the impact of multiple positives in CL, we propose a novel approach, named \MN{}, to effectively utilize multiple positive instances to improve the learning of multi-lingual sentence embeddings. Experimental results on various backbone models and downstream tasks demonstrate that \MN{} leads to better retrieval, semantic similarity, and classification performances compared to conventional CL. We also observe that in unseen languages, sentence embedding models trained on multiple positives show better cross-lingual transfer performance than models trained on a single positive instance.
\end{abstract}

\section{Introduction}
Multi-lingual sentence embedding transforms sentences in different languages into a shared embedding space~\cite{feng2020language, wang-etal-2022-english}, where sentences with similar meanings are positioned close to each other. This is a fundamental and important task in Natural Language Processing (NLP), with various applications including multi-lingual retrieval~\cite{yang2020multilingual}, cross-lingual classifications~\cite{hirota2020emu}, and multi-lingual inference~\cite{Conneau2018xnli}.

As contrastive learning (CL) exhibits great strength on learning sentence representation, CL-based methods have become the common practice
for learning monolingual ~\cite{simcse, su2022one, ni-etal-2022-sentence} as well as multi-lingual sentence embeddings~\cite{feng2020language, wang-etal-2022-english}.
Typically, conventional CL performs with an anchor, a positive and multiple negative examples. The learning objective of CL is to pull the anchor and the positive closer and push the anchor and the negatives apart~\cite{Oord2018RepresentationLW}. 

\begin{figure}[]
    \centering
    \subfigure[Mono-lingual]{
        \includegraphics[width=.16\textwidth]{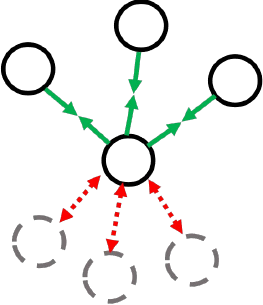}
        \label{fig:intro1}
    }
    \subfigure[Multi-lingual]{
        \includegraphics[width=.18\textwidth]{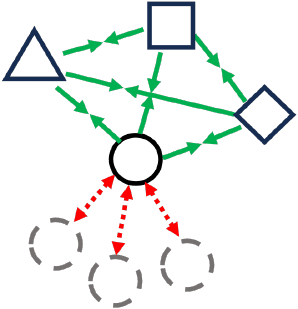}
        \label{fig:intro3}
    }
    \subfigure[Mono-lingual]{
        \includegraphics[width=.218\textwidth]{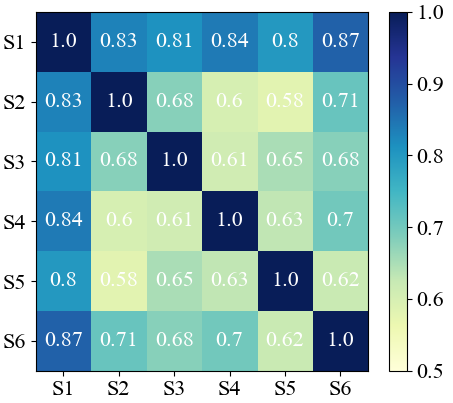}
        \label{fig:intro2}
    }
    \subfigure[Multi-lingual]{
        \includegraphics[width=.22\textwidth]{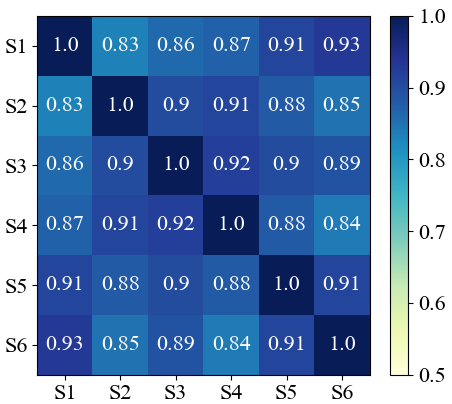}
        \label{fig:intro4}
    }
    
    \caption{
    \blue{Different shapes denote examples in different languages. Solid and dotted arrows denote positive and negative pairs, respectively.
    (a) vs. (b): Multi-lingual positives by translation exhibit transitive similarity, while monolingual positives do not.
    (c) vs. (d): Pairwise semantic similarity scores of sampled sentences for mono- and multi-lingual, highlighting the similarity transitive similarity. Example sentences are sourced from XNLI dataset,} refer to \ref{appendix1} for details.
    }
    \label{fig:intro}
\end{figure}

Our work aims at improving contrastive learning for multi-lingual sentence embedding with \emph{multiple positives}.
While existing approaches in multi-lingual sentence embedding only takes the naive CL with a single positive example, we argue that leveraging multiple positives should be considered especially for multi-lingual sentence embeddings. 
In contrast to mono-lingual CL, richer and more complex relationships exist among multiple positives in multi-lingual CL: \textbf{(1) positives in a diverse set of languages}, which can benefit cross-lingual learning; \textbf{(2) transitive similarity across multiple positives by translation}, which provides reliable structural information for learning.


Specifically, 
\blue{we calculate similarity scores among multiple positives to emphasize the unique effect of multiple positives especially in multi-lingual scenarios.} 
As shown in Figure~\ref{fig:intro1} and~\ref{fig:intro2}, although multiple positive examples all share high similarity with the anchor, the transitive similarity does not always exist among positives, e.g., (S1, S2) and (S1, S3) have similar meanings but (S2, S3) do not. 
While in multi-lingual settings shown in Figure~\ref{fig:intro3} and~\ref{fig:intro4} where translations are used as positives, multiple positives can provide cross-lingual information from diverse languages. Moreover, multilingual translation guarantees transitivity of similarity across positive examples, leading to effective CL with multiple positives.

Motivated by the aforementioned discussion, in this paper, we investigate the impact of CL with multiple positives especially for multi-lingual sentence embeddings and propose \MN{} (\textbf{M}ulti-lingual \textbf{P}ositives in \textbf{C}ontrastive \textbf{L}earning), a novel approach for sentence embedding to effectively leverage multiple positives to improve the quality of multi-lingual sentence embeddings.
Specifically, we construct multiple positive instances by collecting multi-lingual translations for the anchor sentence. Besides, we propose to utilize a multi-positive loss function to effectively learn from the multiple positives, in which the conventional contrastive correlation and structural information among multi-lingual translations are learned simultaneously. To the best of our knowledge, we are the first to explore the impact from multiple positives to multi-lingual sentence embeddings.

Extensive experiments on various models and downstream tasks are conducted to evaluate the proposed approach.
Experimental results confirm that leveraging multiple positives leads to better semantic similarity, retrieval, and classification performances on LaBSE~\cite{feng2020language}, with an improvement of 4.1 on the BUCC task, 2.7 on the STS17 task, and 1.5 on the MTOP domain classification task, respectively. The improvement holds for a diverse set of backbone models, including the continual training on well-trained sentence embedding models such as mSimCSE~\cite{wang-etal-2022-english}, as well as training from scratch on pre-trained language models such as mBERT~\cite{devlin-etal-2019bert} and XLM-RoBERTa~\cite{conneau-etal-2020-unsupervised}. \blue{A variant of \MN{} outperforms the state-of-the-art model, mSimCSE, in various evaluation tasks.} We also observe better cross-lingual transfer performances on unseen languages with our proposed \MN{} compared to conventional CL with a single positive instance. Moreover, to investigate the effectiveness of \MN{}, we evaluate variants of \MN{} by incorporating different languages and adjusting the number of languages included in our training dataset.

\section{Multiple Positives in Contrastive Learning for Multi-lingual Sentence Embeddings}
This section begins with a formal definition of learning sentence embeddings with CL, followed by the construction of training data with multiple positives and the utilization of multiple positives for sentence embeddings.

\begin{figure*}
    \centering
    \includegraphics[width=1\textwidth]{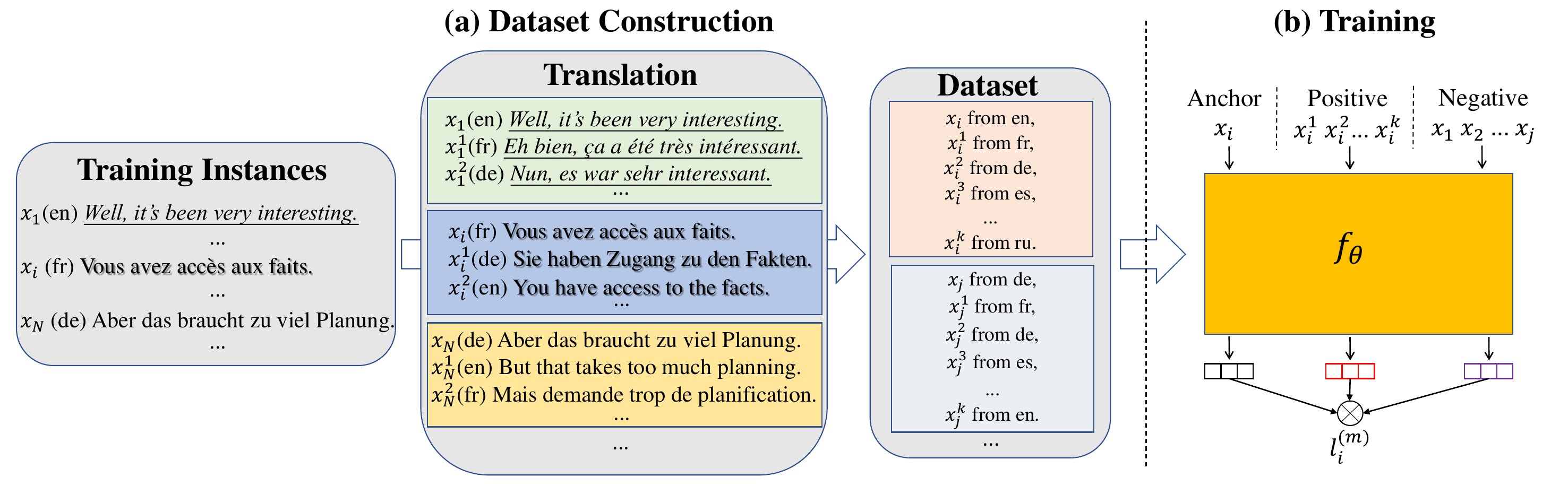}
    \caption{Illustration of \MN{}. Left: we reorganize multi-lingual data with a translation dataset to construct a training dataset with multiple positives. \blue{Sentences in the same font are translations from different languages.} Right: we perform contrastive loss with multiple positive instances to update the model.}
    \label{fig:structure}
\end{figure*}
\paragraph{Contrastive learning for sentence embeddings}
Given a sentence $x_i \sim \blue{\mathbb{X}}$, 
sentence embedding learning aims to learn a parameterized network $f_{\theta}$. The network can be applied to $x_i$ to obtain a dense vector, i.e., $\mathbf{h}_i = f_{\theta}(x_i)\in\mathbb{R}^d$, which can represent the semantic meaning of sentence $x_i$.
The idea of contrastive learning is to construct a positive example $x_i^+$ for $x_i$ and pull them close, while keeping $x_i$ far from other negative examples.
A commonly used training objective~\cite{Oord2018RepresentationLW, simcse, wu2022pcl} is to minimize the following contrastive loss:
\begin{equation}
    l_i^{(s)} = -\log\frac{e^{sim(\mathbf{h}_{i}, \mathbf{h}_{i}^{+})/\tau}}
    {\sum_{j=1}^{N} e^{sim(\mathbf{h}_{i}, \mathbf{h}_{j})/\tau}},
    \label{eq:single}
\end{equation}
where $\mathbf{h}_i^+ = f_{\theta}(x_i^+)$, $sim(\cdot, \cdot)$ is a similarity metric, $N$ is the size of a mini-batch, and $\tau$ is the temperature parameter. 
After training, these semantically meaningful embeddings can be used to represent sentences for various downstream tasks such as sentence retrieval, sentence-level classification, and semantic textual similarity.

\paragraph{Multiple Positives in Contrastive Learning}
The conventional approach cannot fully capture the semantic richness and diverse expressions in different languages. To address this limitation, in this work, we propose to leverage multiple positives in contrastive learning to improve the learning of multi-lingual sentence embeddings.
Unlike the conventional single-positive CL loss in equation~\ref{eq:single}, \blue{for anchor sentence $x_i$, we construct a multiple positives set from multi-lingual translation $ X_i^{mp} = \{x_i^1, ..., x_i^K\} $} 
, where $K$ is the number of positives from different languages.
Inspired by previous works that deal with multiple positives~\cite{pmlr-v97-frosst19a, NEURIPS2020_d89a66c7}, the training objective of CL with multiple positives in multi-lingual sentence embedding is shown as follows:  
\begin{equation}
    l_i^{(m)} = -\log\frac{\sum_{k=1}^{K}e^{sim(\mathbf{h}_i, \mathbf{h}_i^k)/\tau}}
    {\sum_{j=1 \land j \neq i}^{N} e^{sim(\mathbf{h}_i, \mathbf{h}_j)/\tau}},
    \label{eq:multi}
\end{equation}
where $\mathbf{h}_i^k$ stands for the representation of positive sentence $x_i^k$ from positive set $X_i^{mp}$, and $N$ is the size of a mini-batch. Equation~\ref{eq:multi} allows us to capture the linguistic diversity and complex relationships among sentences across different languages.

\paragraph{Dataset Construction in \MN{}}
Figure~\ref{fig:structure} illustrates the dataset construction in \MN{}. In order to utilize the transitivity among positives, we collect translations as multiple positives \blue{$X_i^{mp}$ from a multi-lingual translation dataset for $x_i$. We reorganize the multi-lingual training instances by assembling translations into one group.} Sentences within the same group share the same meaning and exhibit transitive similarity, allowing us to have one anchor sentence $x_i$ and a positive set $X_i^{mp}$ to perform the \MN{} loss in Equation~\ref{eq:multi}.

\section{Experiments}

\subsection{Details of Training Dataset}
The multi-lingual translation dataset used in our experiment is the XNLI~\cite{Conneau2018xnli} dataset. \blue{Considering the intersection of different evaluation tasks, six languages English~(\textit{en}), German~(\textit{de}), French~(\textit{fr}), Spanish~(\textit{es}), Russian~(\textit{ru}), and Chinese~(\textit{zh}), are selected in our dataset so that we can evaluate the effects on both seen and unseen languages simultaneously with minimal influence from other languages. Other combinations of languages will be discussed in Section~\ref{3.7}. 
Sentences in languages other than English are derived from translations given in XNLI. This allows us to assemble multi-lingual translations into the same group. Specifically, when dealing with a given sentence, we exclusively choose the sentence itself, omitting its corresponding entailment, neutral, and contradictory counterparts provided in XNLI.}


Finally, our dataset comprises 400k data groups. In this dataset, for each anchor sentence, we can access multiple positives at the same time. \blue{Note that each language has the same probability of serving as the anchor sentence.} We perform a wide range of experiments with various models on this dataset to verify the effects of multiple positives. 

\subsection{Baselines}
Several strong baselines are chosen for comparison. The first selections are two state-of-the-art multi-lingual sentence embedding models trained on one single positive instance, LaBSE~\cite{feng2020language} and mSimCSE~\cite{wang-etal-2022-english}. We specifically choose $mSimCSE_{all}$, a variant of mSimCSE that includes 15 languages during training and utilizes hard-negative examples.

In addition, we select some general models such as Sentence-T5~\cite{ni-etal-2022-sentence}, LASER, and LASER2~\cite{artetxe2019massively} to compare performance on different tasks. Besides, \texttt{bert-base-multilingual-uncased}\footnote{\url{https://huggingface.co/bert-base-multilingual-uncased}}~\cite{DBLP:conf/naacl/DevlinCLT19} and \texttt{xlm-roberta-large}\footnote{\url{https://huggingface.co/xlm-roberta-large}}~\cite{conneau-etal-2020-unsupervised} are included as alternative Pretrained Language Models (PLMs).

\subsection{Training Details}
We continuously train various backbone models including LM base models and sentence embedding base models. The training of all our models is conducted on one NVIDIA A100 80G.
The batch size is set to 128, the maximum sequence length is set to 64, and the learning rate is 1e-5.
Particularly for LM base model mBERT and XLM-RoBERTa, the conventional contrastive loss in Equation~\ref{eq:single} is initially used for warm-up, during which the learning rate is set to 2e-5 for 2000 steps.

The temperature parameter $\tau$ is set to 0.05. We use \texttt{[cls]} token as sentence embedding. Cosine similarity is used as the similarity metric, which allows us to compute the similarity distribution by contrasting anchor sentence with multiple positives and negatives. Accounting for the margin among multiple positives and negative examples, we specifically use min-max scaling to rescale the similarity scores within a range of $[-1/\tau, 1/\tau]$.
Under our specified training details, BERT-base size models require approximately 30G of memory, while BERT-large size models require about 60G of memory.

We evaluate the models on development sets every 125 steps to find the best checkpoints. Specifically, STS22 and STS17 are used as the development set for each other. We also use Tatoeba and BUCC as each other's development sets for bi-text mining tasks. For classification tasks, we directly use the validation set provided by MTOP domain classification as development set. All of our results are obtained from an average of five random seeds.

\subsection{Evaluation Tasks}
We evaluate models on three fundamental multi/cross-lingual tasks: bitext mining, semantic similarity and classification. We run semantic similarity and classification with MTEB\footnote{\url{https://huggingface.co/spaces/mteb/leaderboard}}~\cite{muennighoff2022mteb}, and bitext-mining with XTREME\footnote{\url{https://github.com/google-research/xtreme}}~\cite{hu2020xtreme} benchmark.

\paragraph{Bitext Mining} is a retrieval task where a sentence and a paragraph (or longer sentence) will be given. The tested model is supposed to find the best match for the sentence in the paragraph by calculating cosine similarity for each pair of embedded sentences. We evaluate our trained models using 14 and 36 Tatoeba~\cite{artetxe2019massively} and BUCC~\cite{zweigenbaum-etal-2017-overview} datasets through the XTREME benchmark. 
We report the f1 score for BUCC and the accuracy for Tatoeba.

\paragraph{Semantic Similarity} requires the models to calculate two given sentences' similarity (scores). Higher scores generally mean higher similarity. We choose the cross-lingual STS17~\cite{cer-etal-2017-semeval} and STS22~\cite{chen-etal-2022-semeval} and report Spearman correlation scores based on cosine similarity metrics. Note that in STS22, all of our averaged results do not contain French-Polish~(\textit{fr-pl}) this particular language pair because we find this pair in MTEB benchmark appears to be unstable, and even totally different models can have exactly the same correlation score on MTEB leaderboard.

\paragraph{Classification} tasks require the model to determine the label of given sentences based on their sentence embeddings. An additional classifier layer will be trained on the given training set, and the performance of the model will be tested on the test set. We choose the MTOP Domain Classification task~\cite{li-etal-2021-mtop} through MTEB benchmark and report the accuracy metric.

\begin{table*}[]
\centering
\resizebox{\textwidth}{!}{
\begin{tabular}{lccccccc}
\hline
\textbf{Model} & \textbf{BUCC} & \textbf{Tatoeba(14avg.)} & \textbf{Tatoeba(36avg.)} & \textbf{STS17} & \textbf{STS22} & \textbf{MTOP} & \blue{\textbf{Avg.}$^{\dagger*}$}\\ \hline
\hline
\multicolumn{8}{c}{\textit{LM Base Models}} \\
\hline
mBERT~\cite{DBLP:conf/naacl/DevlinCLT19}            & 56.7  &-    & -    & - & -   & - & -\\
\ \ \ \ + Single  & 84.1$_{\pm0.09}$ & 70.5$_{\pm0.34}$ & 64.4$_{\pm0.45}$ & 57.0$_{\pm0.26}$ & 53.4$_{\pm0.63}$ & 62.3$_{\pm0.21}$ & 65.3\\
\ \ \ \ + \textbf{Multiple (Ours)}  & \textbf{85.3$_{\pm0.36}$~($\uparrow$1.2)} & \textbf{71.6$_{\pm0.20}$~($\uparrow$1.1)} & \textbf{65.1$_{\pm0.34}$~($\uparrow$0.7)} & \textbf{57.8$_{\pm0.31}$~($\uparrow$0.8)} & \textbf{55.8$_{\pm0.56}$~($\uparrow$2.4)} & \textbf{62.7$_{\pm0.20}$~($\uparrow$0.4)} & 66.4\\
XLM-R~\cite{conneau-etal-2020-unsupervised} & 66.0 & 57.6 & 53.4 & - & - & - & - \\
\ \ \ \ + Single & 94.5$_{\pm0.17}$  & 91.3$_{\pm0.08}$  & 89.6$_{\pm0.19}$ & 71.1$_{\pm0.63}$ & 59.8$_{\pm0.44}$ & 83.0$_{\pm0.19}$ & 81.6\\
\ \ \ \ + \textbf{Multiple (Ours)}  & \textbf{95.7$_{\pm0.36}$~($\uparrow$1.2)} & \textbf{92.0$_{\pm0.11}$~($\uparrow$0.8)} & \textbf{90.4$_{\pm0.13}$~($\uparrow$0.8)} & \textbf{73.2$_{\pm0.20}$~($\uparrow$2.1)} & \textbf{61.4$_{\pm0.41}$~($\uparrow$1.6)}  & \textbf{84.5$_{\pm0.39}$~($\uparrow$1.5)} & 82.9\\
\blue{XLM-R w/ hard negative} &-&-&-&-&-&-&-\\
\ \ \ \ + Single (mSimCSE$_{all}$)~\cite{wang-etal-2022-english} & 95.2 & 93.2 & 91.4 & 76.7 & 63.2 & 84.1 & 84.0\\
\ \ \ \ + \blue{\textbf{Multiple (Ours)}$^{\dagger}$} & \textbf{95.4$_{\pm0.27}$~($\uparrow$0.2)} & \textbf{93.5$_{\pm0.13}$~($\uparrow$0.3)} & \textbf{91.8$_{\pm0.07}$~($\uparrow$0.4)} & \textbf{78.9$_{\pm0.47}$~($\uparrow$2.2)} & \textbf{64.0$_{\pm0.16}$~($\uparrow$0.8)} & \textbf{86.8$_{\pm0.30}$~($\uparrow$2.7)} & 85.1\\
\hline
\hline
\multicolumn{8}{c}{\textit{Sentence Embedding Base Models}} \\
\hline
INFOXLM~\cite{chi-etal-2021-infoxlm} & -   & 77.8          & 67.3 & - & - &-  & -\\
LASER~\cite{artetxe2019massively}            & 92.9  & 95.3 & 84.4 & -    & -     & -    & -\\ 
LASER2~\cite{artetxe2019massively}          & -    & -   & -    & 69.2 & 41.6  & 73.5 & -\\ 
Sentence-T5-large~\cite{ni-etal-2022-sentence} & -   & -  & -    & 44.4 & 47.0  & 61.5 & -\\ 
mSimCSE$_{all}$~\cite{wang-etal-2022-english}  & 95.2 & 93.2 & 91.4 & 76.7  & 63.2 & 84.1   & 84.0       \\
\ \ \ \ + \textbf{Multiple (Ours)} & 96.0$_{\pm0.23}$~($\uparrow$0.8) & 93.5$_{\pm0.10}$~($\uparrow$0.3) & 91.4$_{\pm0.14}$ & \textbf{78.5$_{\pm0.23}$~($\uparrow$1.8)}  & \textbf{64.3$_{\pm0.28}$~($\uparrow$1.1)} & 85.9$_{\pm0.27}$~($\uparrow$1.8)  & 84.9      \\
LaBSE~\cite{feng2020language}   & 93.5 & 95.3 & 95.0 & 74.2 & 60.9  & 84.6 & 83.9\\
\ \ \ \ + \textbf{Multiple (Ours)} & \textbf{97.6$_{\pm0.19}$~($\uparrow$4.1)}  & \textbf{96.0$_{\pm0.08}$~($\uparrow$0.7)} & \textbf{95.4$_{\pm0.07}$~($\uparrow$0.4)} & 76.9$_{\pm0.22}$~($\uparrow$2.7) & 61.5$_{\pm0.29}$~($\uparrow$0.6) & \textbf{86.1$_{\pm0.26}$~($\uparrow$1.5)} & 85.6 \\
\hline
\end{tabular}}
\caption{Overall results of different models on various downstream tasks. \blue{We report the average scores and their corresponding standard deviation from five random seeds for each task.} We adopt the baseline's results from \citet{hu2020xtreme} and \citet{muennighoff2022mteb}. +Single stands for models we continually train on parallelized dataset with one single positive while +Multiple stands for models we train on our proposed method with multiple positives. \blue{$^{\dagger}$: This variant is trained with all 15 languages in XNLI and combined with hard negatives. $^{\dagger*}$: This column refers to the average score of all six tasks, showing statistically significant results with p-value < 0.005 when comparing each +Multiple to its corresponding base model.} 
}
\label{table_overall}
\end{table*}

\subsection{Main Experimental Results}
In this section, we present the main experimental results. In particular, +~Multiple refers to models that are trained through our proposed framework with five positive instances. To facilitate a fair comparison with conventional CL with one single positive, we modify our dataset to follow a parallel structure, where only source-target pairs from different languages are included. \blue{For example, in our main experiments, we have six multi-lingual sentences in one group so this group will be converted into three random language pairs in the parallel dataset.} +~Single refers to the models that are trained on this modified dataset. Note that the data in this modified dataset are the same as those in the original dataset, with the only difference being their structure \blue{and the model possesses an equal chance to see all sentences for one time in both datasets. Unless otherwise specified, both +~Multiple and +~Single refer to models trained without hard negatives.}

More specifically, we aim to address two questions based on the overall results of various downstream tasks shown in Table~\ref{table_overall}.
\textbf{Q1}: Does the utilization of multiple positives yield more substantial benefits to the model compared to conventional CL with a single positive?
\textbf{Q2}: Does the effectiveness of leveraging multiple positives still hold for stronger sentence embedding models? \blue{Refer to Appendix~\ref{appendix2} for detailed results of different models.}

\subsubsection{Multiple Positives Yield Better Performance than Single Positive}

We first train general LMs with a warm-up under our proposed methods and compare their performance with LMs trained on conventional CL. 

From the upper part of Table~\ref{table_overall}, it is evident that continuous training with both single and multiple positives on pretrained language models enhances the performance of the models. However, from the average score of the downstream tasks, we can observe that models trained using our proposed framework with multiple positives demonstrate stronger performance than conventional CL-based methods with single positives. Specifically, we discover an average improvement of 1.1 for mBERT and 1.3 for XLM-R across different downstream tasks compared to models trained on single positive. The most significant improvement is seen in STS22 for mBERT, and in STS17 for XLM-R.

In order to fairly compare MPCL with state-of-the-art models, we train a variant with the same language coverage and apply hard negatives~\cite{kalantidis2020hard}. We include all 15 languages in XNLI and utilize sentences with contradictory labels as hard negatives. This comparison is referred as \textbf{XLM-R w/ hard negative} shown in Table~\ref{table_overall}. Our XLM-R trained with multiple positives surpasses mSimCSE$_all$, which is trained with single positives in all tasks we report. Note that during the training of this variant, an anchor sentence can access five positives from different languages and the hard negative instance is also randomly selected from all languages. This variant indicates that our proposed MPCL can be incorporated with other existing orthogonal methods, such as hard negatives, and contributes to better performance.

These observations suggest that for general multi-lingual language models, leveraging multiple positives can offer a richer and more useful source of information for training, thus yielding more substantial benefits to the model. Note that with the exception of the BUCC task, all the other evaluation tasks include languages that are excluded from our dataset.

\subsubsection{Multiple Positives Can Further Improve Sentence Embedding Models}
Next, we continually train sentence embedding models, including LaBSE and mSimCSE to validate the effectiveness of multiple positives for pretrained sentence embedding models. For mSimCSE$_{all}$, their results of BUCC and Tatoeba are adopted from \citet{wang-etal-2022-english} while we use its released checkpoint~\footnote{\url{https://github.com/yaushian/mSimCSE}} to evaluate its performance on STS17, STS22, and MTOP Domain Classification. LaBSE's results are adopted from the MTEB leaderboard. 

Upon examining the lower part of Table~\ref{table_overall}, we can observe that the highest scores for each task are obtained by models continually trained on our framework. 
At the same time, it is evident that the improvements in the BUCC, STS, and MTOP Domain Classification tasks are more prominent compared to the Tatoeba dataset. A possible reason for this observation is that we only include five languages, excluding English in our dataset, in contrast to the comprehensive evaluation of 36 languages in the Tatoeba dataset. These findings imply that the utilization of multiple positives can even significantly enhance the performance of well-trained sentence embedding models, indicating the robustness of our proposed method.

\begin{figure*}[t]
    \centering
    
    \subfigure[Performance variation on LaBSE]{
        \includegraphics[width=.485\textwidth]{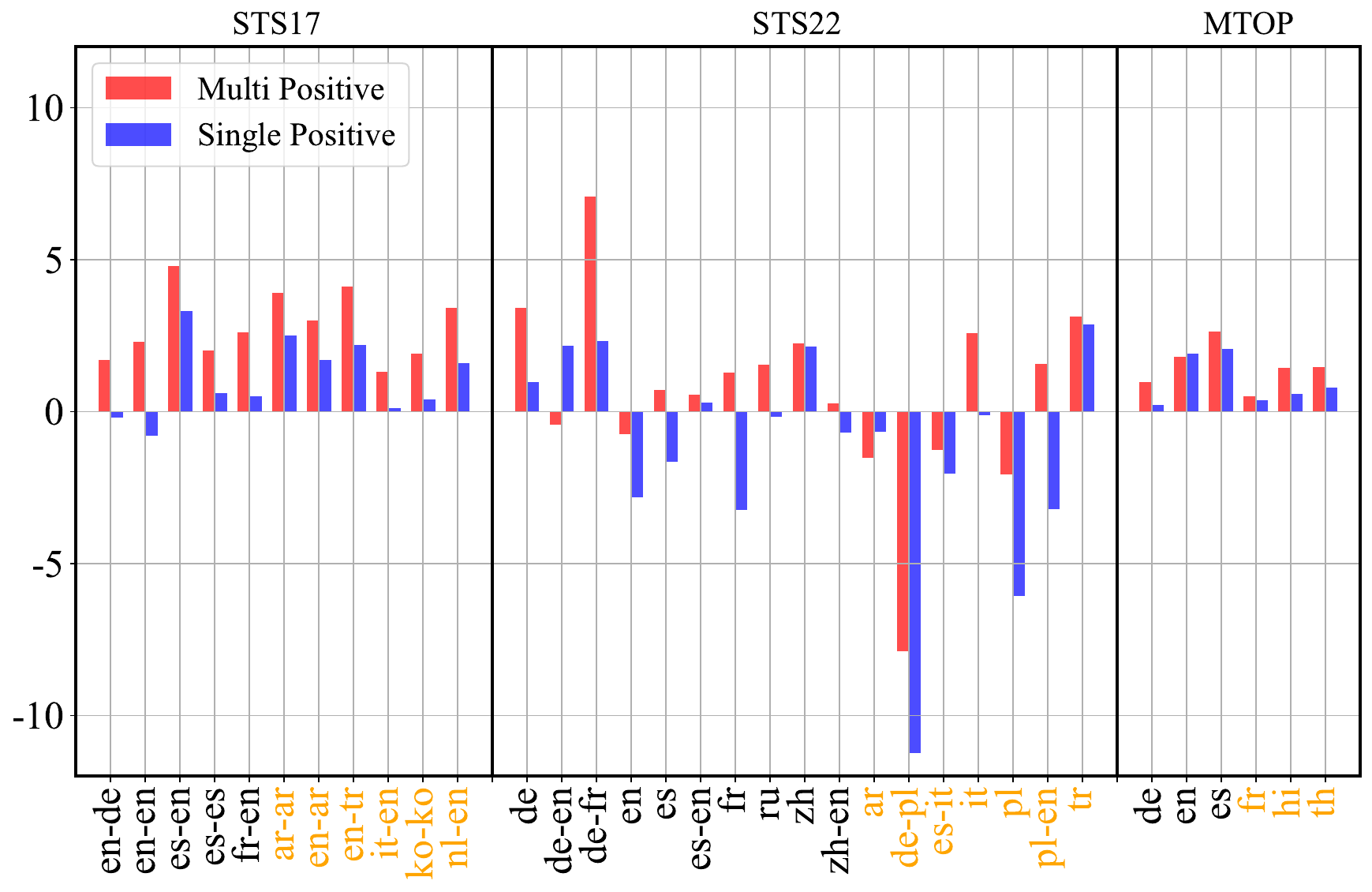}
        
    }
    \subfigure[Performance variation on mSimCSE$_{all}$]{
        \includegraphics[width=.485\textwidth]{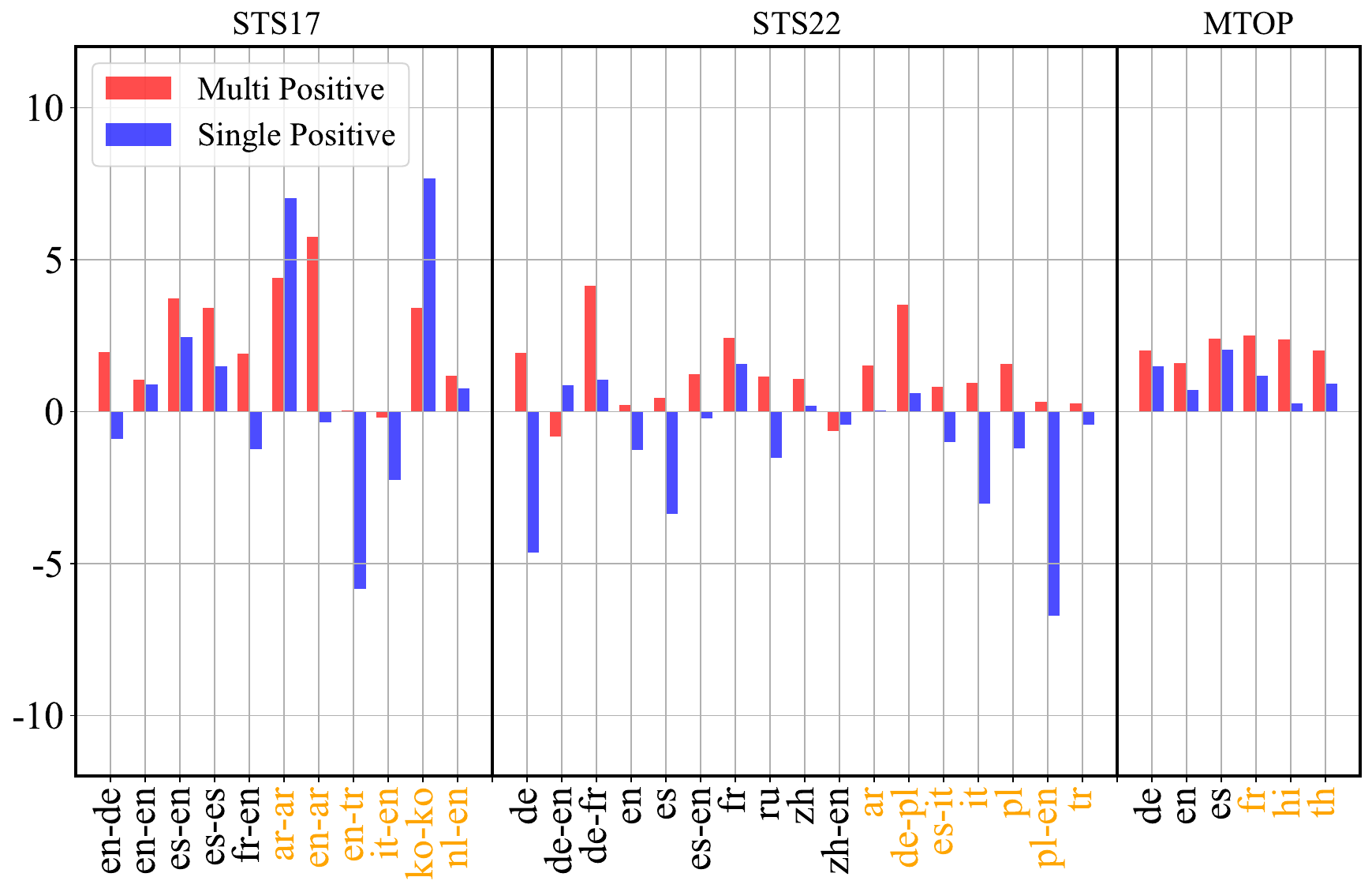}
        
    }
    
    \caption{Detailed performance changes of LaBSE and mSimCSE$_{all}$ on different languages after training on multiple positive or single positive. On the horizontal axis, 0 represents the original scores without additional training. The vertical axis shows changes in performance after further training. Bars above the horizontal axis indicate improvements, while those below indicate decreases. The bold lines split the results into three different parts: STS17, STS22 and MTOP Domain Classification from left to right. Orange color highlights unseen languages.}
    
    \label{fig:change}
\end{figure*}
\subsection{Transferring to Unseen Languages}
Despite the presence of unseen languages in all of our downstream evaluation tasks, models trained using multiple positives consistently demonstrate improvements across almost all overall results. This motivates us to delve deeper into the transferability of multiple positives. We show the results of all language pairs from STS17, STS22 and MTOP Domain Classification in Figure~\ref{fig:change}.

The value of 0 on horizontal axis stands for the scores of the origianl LaBSE and mSimCSE$_{all}$.
For example, in Figure~\ref{fig:change}, the first pair English-German~(\textit{en-de}) indicates that parallel training with a single positive leads to a slight decrease in performance for both LaBSE and mSimCSE, while training with multiple positives leads to an improvement. Note that our training dataset only contains \textit{en}, \textit{de}, \textit{fr}, \textit{es}, \textit{ru} and \textit{zh} so all the other language pairs are considered unseen. Language pairs such as English-Turkish~(\textit{en-tr}), where only one of the two languages is observed in the training set are also considered as exclusive. 

In Figure~\ref{fig:change}, we can observe that models trained with multiple positives generally exhibit better transfer ability to unseen languages. For example, when looking at the unseen language pair \textit{pl-en} in STS22, we observe a drop in accuracy with single positive training, while multiple positive training still shows a significant improvement in LaBSE. This trend can also be observed in mSimCSE's variance, indicating that the transfer ability of multiple positives surpasses that of single positive training. Average results of unseen languages in STS tasks can be found in Appendix~\ref{appendix2}, Table~\ref{table_sts}.

\subsection{The Choice of Languages in Training Set.}
\label{3.7}
In our main experiments, we choose six different languages to satisfy various downstream tasks. However, the composition of the dataset, including the number and selection of positives, still remains unexplored. In this section, we alter the composition of training datasets and specifically choose XLM-R to assess different downstream tasks.

\subsubsection{Training with Only Unseen Language}
\begin{figure}[t]   
    \centering
    \includegraphics[width=.47\textwidth]{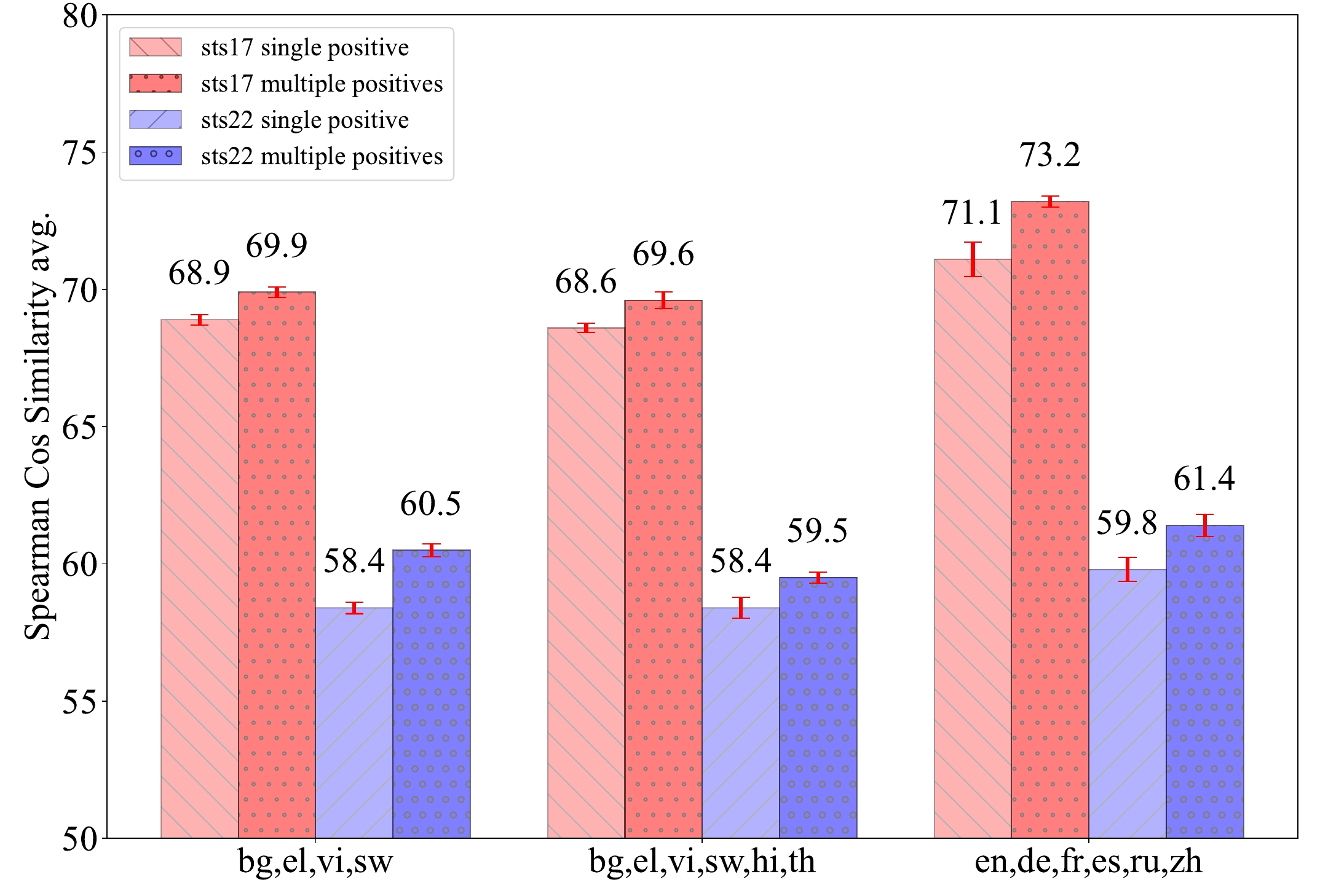}
    \caption{Results of the STS17 and STS22 tasks trained on fully none-overlapping languages. We report the average scores of all language pairs included in STS17 and STS22. \blue{Error bars refer to standard deviation.}}
   \label{fig:transfer}
   
\end{figure}

We first construct another training dataset using multiple positives extracted from XNLI which comprises only four languages: Bulgarian~(\textit{bg}), Greek~(\textit{el}), 
Vietnamese~(\textit{vi}) and 
Swahili~(\textit{sw}) that do not overlap with any language that will be tested in STS17 and STS22 so that all results exhibit models' fully transfer abilities. Besides, we add two more languages: Hindi~(\textit{hi}) and 
Thai~(\textit{th}) to see whether cross-lingual signals from more non-overlap languages can help improve the transfer ability. Figure~\ref{fig:transfer} shows the results.

As reported in \citet{wang-etal-2022-english}, we also observe that CL-based methods exhibit remarkable transfer abilities on totally unseen languages. From Figure~\ref{fig:transfer}, we can observe a slight drop when incorporating two additional languages into the dataset.
This finding aligns with the observation presented by~\citet{conneau-etal-2020-unsupervised}, where they highlight a trade-off between the number of languages and transfer performance. However, the strong transfer ability of multiple positives remains evident as training on completely non-overlapping languages can still yield competitive results compared to our original training dataset. Besides, we also observe an obvious trend in Figure~\ref{fig:transfer} where all average results from multiple positives consistently surpass those from single positives. We believe that by bringing anchor sentences and all the remaining positives closer together, models can effectively capture the nuances of different languages and achieve better language representation.

\subsubsection{Training with Only Seen Language}
We also explore the effect of training with only languages that will be tested in the STS17 and STS22 tasks. We remove the \textit{ru} and \textit{zh} from our dataset, which will not be evaluated in STS17 to see whether \textit{ru} and \textit{zh} provide more useful cross-lingual signals. Besides, we add two more additional languages, 
Arabic~(\textit{ar}), and Turkish~(\textit{tr}) so that all the languages in this dataset~(\textit{en}, \textit{de}, \textit{fr}, \textit{es}, \textit{ar}, \textit{tr}) will be tested in the STS17 and STS22 tasks. The results are shown in Table~\ref{table_allin}.

\begin{table*}[t]
\centering
\resizebox{\textwidth}{!}{
\begin{tabular}{lccccccccc}
\hline
\textbf{Model} & \textbf{en-ar} & \textbf{en-tr} &\textbf{STS17 avg.} & \textbf{ar} & \textbf{ru} &  \textbf{tr} & \textbf{zh} & \textbf{zh-en} & \textbf{STS22 avg.}\\ \hline 
XLM-R \\
\ \ \ \ +Single & 67.9$_{\pm0.49}$ & 68.4$_{\pm0.71}$ & 71.1$_{\pm0.63}$ & 57.2$_{\pm0.64}$ & 59.1$_{\pm0.39}$ & 64.4$_{\pm0.43}$ & 65.6$_{\pm0.25}$ & 67.4$_{\pm0.65}$ & 59.8$_{\pm0.44}$ \\
\ \ \ \ +Multiple & 71.0$_{\pm0.24}$ & 72.5$_{\pm0.40}$ & \textbf{73.2}$_{\pm0.20}$ & 57.1$_{\pm0.48}$ & \textbf{60.3$_{\pm0.20}$} & 64.3$_{\pm0.23}$ & 64.8$_{\pm0.27}$ & \textbf{68.6$_{\pm0.22}$} & \textbf{61.4}$_{\pm0.41}$ \\
\ \ \ \ +Single, -\textit{ru, zh} & 67.6$_{\pm0.18}$    & 70.3$_{\pm0.88}$ & 70.6$_{\pm0.64}$ & 55.6$_{\pm0.46}$ & 57.5$_{\pm0.28}$ & 63.1$_{\pm0.60}$ & 65.3$_{\pm0.17}$ & 67.3$_{\pm0.42}$ & 59.6$_{\pm0.46}$      \\
\ \ \ \ +Multiple, -\textit{ru, zh} & 70.3$_{\pm0.49}$  & 70.8$_{\pm0.23}$ & 71.8$_{\pm0.48}$  & 56.2$_{\pm0.07}$ & 58.4$_{\pm0.67}$ & 62.7$_{\pm0.60}$ & 65.7$_{\pm0.28}$ & 68.1$_{\pm0.32}$ & 60.2$_{\pm0.31}$  \\
\ \ \ \ +Single, +\textit{ar, tr} & 71.2$_{\pm0.61}$  & 72.1$_{\pm0.58}$ & 71.3$_{\pm0.44}$ & 58.7$_{\pm0.33}$ & 58.3$_{\pm0.32}$ & \textbf{65.7}$_{\pm0.26}$ & 64.8$_{\pm0.51}$ & 67.0$_{\pm0.37}$ & 60.3$_{\pm0.36}$  \\
\ \ \ \ +Multiple, +\textit{ar, tr}  & \textbf{72.5}$_{\pm0.18}$       & \textbf{73.6}$_{\pm0.49}$ & 72.1$_{\pm0.27}$ & \textbf{59.0}$_{\pm0.27}$ & 59.3$_{\pm0.61}$ & 65.3$_{\pm0.41}$ & \textbf{66.3}$_{\pm0.34}$ & 67.3$_{\pm0.35}$ & 60.9$_{\pm0.34}$\\ 
\hline
\end{tabular}
}
\caption{\blue{Results of different composition of dataset on STS17 and STS22 task.} -\textit{ru, zh} means we only have four languages in the dataset while +\textit{ar, tr} means that we replace \textit{ru} and \textit{zh} with \textit{ar} and \textit{tr}. }
\label{table_allin}
\end{table*}

We focus specifically on the language pairs that we removed or added such as English-Arabic~(\textit{en-ar}) and English-Turkish~(\textit{en-tr}) to examine the impact of different dataset compositions. Upon removing \textit{ru} and \textit{zh} from the original dataset, we observe a slight decrease in the overall accuracy of STS17. This suggests that \textit{ru} and \textit{zh} in fact contribute valuable cross-lingual signals to the model in STS17. However, in STS22, although the performance of \textit{ru} drops due to its removal, the accuracy of \textit{zh} acquires an interesting improvement. One possible reason for this may be that although we include \textit{zh} in our dataset, we do not have the exact same \textit{zh-zh} pairs that are tested in STS22. Thus, the model is able to learn information from other cross-lingual signals. 

In an attempt to replace \textit{ru} and \textit{zh} with \textit{ar} and \textit{tr}, we observe significant improvements in the \textit{ar} and \textit{tr} related pairs in both STS17 and STS22, but this change leads to a decline in the overall accuracy compared to our original results.  \citet{dhamecha-etal-2021-role} have noted that languages with high relatedness can mutually benefit each other. Therefore, adding languages like \textit{ar} which has low linguistic relatedness with other languages may have an impact on the performance of other languages.

\subsection{Case study}
\begin{figure}[H]
    \centering
        \subfigure[``We had a great talk.'' ]{
        \includegraphics[width=.35\textwidth]{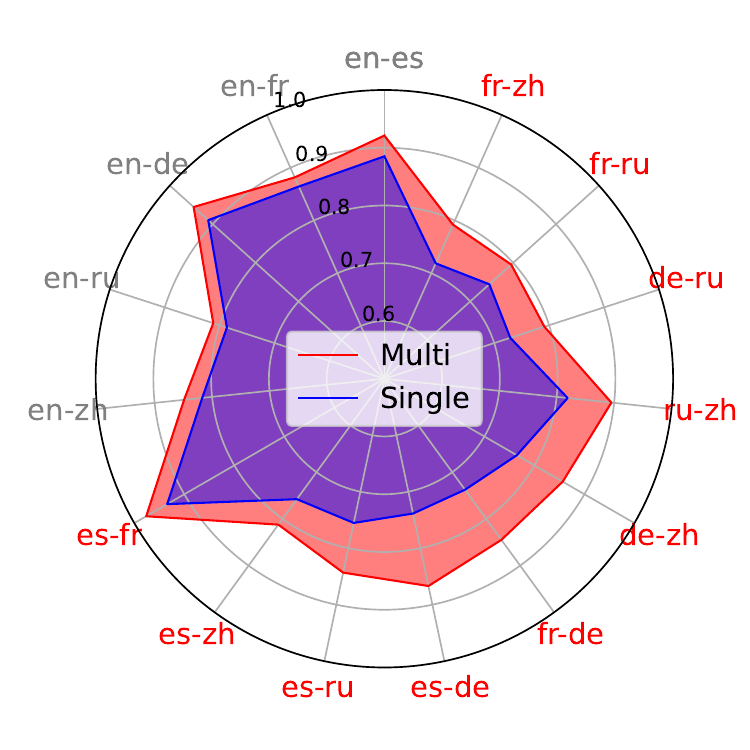}
       
    }
    \subfigure[``I don't know whether he stayed in Augusta after that.'' ]{
        \includegraphics[width=.35\textwidth]{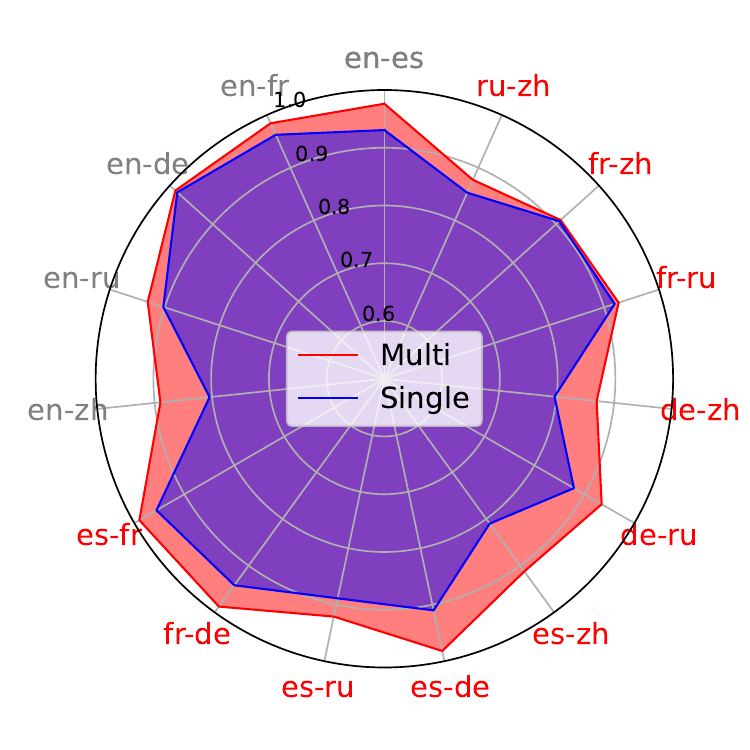}
        
    }

    \caption{Cross-lingual similarity scores calculated by XLM-R + Multiple or XLM-R + Single for two randomly chosen examples. The example sentences are shown in the sub-figure caption. Non-English-centric language pairs are highlighted with red color.}
    \label{case}

\end{figure}

In this section, we randomly choose two examples from XNLI test set to demonstrate effect of multiple positives on cross-lingual similarity. With the six languages included in our experiment, we can obtain fifteen cross-lingual pairs. The similarities are calculated for all fifteen language pairs. The results are shown in Figure~\ref{case}. The sub-figure caption is the example sentence in English. Non-English-centric pairs are highlighted with red color. As we calculate similarity scores based on translations, the gold label similarity score for all cross-lingual pairs should be 1.0. 
As shown in the figure, it can be observed that for most English-centric pairs, single positive and multiple positive models achieve a comparable level of similarity. However, for non-English-centric pairs, multiple positive model exhibits an obvious higher similarity. This indicates that our approach of utilizing translations as multiple positives improves the cross-lingual representation learning, especially for non-English-centric language pairs.

\section{Related Work}
 
\subsection{Contrastive Learning}
Conventional contrastive learning performs with an anchor, one positive instance and multiple negative instances by pulling the distance between positive instances closer and between negative instances farther.
The idea of contrastive loss can be traced back to \citet{Chopra2005LearningAS}. Later, the wide usage of contrastive loss in the field of computer vision takes it to a higher level and has been proven to be an effective way of learning representations~\cite{Wu2018UnsupervisedFL, He2019MomentumCF, chen2020simple}.  
In the field of NLP, contrastive learning has been applied into a variety of tasks such as machine translation~\cite{pan2021contrastive}, text classification~\cite{du2021constructing}, summarization~\cite{duan2019contrastive}. Recently, it has also been shown that CL plays a significant role in cross-modal representation learning~\cite{li-etal-2021-unimo, radford2021learning} which indicates that even pulling positive instances from different modalities can be beneficial. 

\blue{Contrastive learning with multiple positives has been studied in previous researches in computer vision~\cite{NEURIPS2020_d89a66c7}, with fine-grained strategies such as soft-nearest neighbor~\cite{pmlr-v97-frosst19a} and ranking~\cite{Dwibedi_2021_ICCV, hoffmann2022ranking}. In this paper, as the focus is verifying the impact of the usage of multiple translated positives for sentence embedding, we simply assign equal importance to all positives.}
\subsection{Mono-lingual Sentence Embeddings}
SimCSE~\cite{simcse} exploits the ability of CL by using dropout as augmentation in unsupervised settings. For supervised sentence embeddings, SimCSE makes use of the NLI dataset~\cite{nlibowman2015large} to establish positive and hard negative examples. The success of SimCSE attracts researchers' attention to CL when dealing with sentence embeddings~\cite{ni-etal-2022-sentence, wang2022text, su2022one, Xie2022StableCL} since CL-based models provide more competitive results on downstream tasks than classical models such as S-BERT (Sentence-BERT)~\cite{reimers-2019-sentence-bert}. \citet{su2022one} further combine CL with prompt-like instructions while \citet{liu-etal-2023-rankcse} leverages ranking information into CL.

For now, few works exist considering multiple positives for mono-lingual sentence embeddings. \citet{wu2022pcl} initially expands a single positive instance into multiple positives for one anchor sentence through multiple augmentations.

\subsection{Multi-lingual Sentence Embeddings}
Training a universal sentence embedding model for all languages is a fundamental and important task. \blue{As the ability of efficient similarity calculation across languages, multi-lingual sentence embeddings have been applied to low-resource corpus filtering~\cite{chaudhary2019low}, parallel corpus mining~\cite{kvapilikova2021unsupervised}, and synthesized dataset filtering~\cite{, wu-etal-2023-wspalign}.}
\citet{artetxe2019massively} utilize the BiLSTM structure with a shared vocabulary for all languages. Recent studies prefer to adopt CL in multilingual settings with source-target translation pairs where the target sentence would be considered as the particular positive for the source sentence.
For example, LaBSE~\cite{feng2020language} is a BERT-based multilingual sentence representation model trained on massive amounts of monolingual data and translation pairs covering over 100 languages. \citet{mao-nakagawa-2023-lealla} distill LaBSE for lightweight mutil-lingual sentence embedding models. \citet{wang-etal-2022-english} show that CL resembling SimCSE can also be applied to multi-lingual settings by using multi-lingual data. Besides, sentence-level CL is often combined with token-level information, for example, token-level reconstruction~\cite{mao2022ems} and token-level alignment~\citet{Li2023DualAlignmentPF} to improve cross-lingual sentence embeddings.
Although CL-based models have become common, leveraging multiple positives for learning multi-lingual sentence embeddings is still unexplored.

\section{Conclusion}

In this work, we propose \MN{}, which improves multi-lingual sentence embeddings by utilizing a set of positive examples, specifically multi-lingual translations, for each anchor sentence. Through the incorporation of multiple positives, \MN{} captures both linguistic diversity and transitive similarities, thereby enriching the embedding space. It updates pairwise similarity distributions into group-wise similarity ones by contrasting the anchor with multiple positives and employs a novel multi-positive loss function that simultaneously learns contrastive correlations and structural information among translations. By doing so, \MN{} can improve performance in semantic similarity, retrieval, and classification tasks while exhibiting better robustness during training. Moreover, it is also observed in the experiments that \MN{} shows a better transfer ability on unseen languages.

\section*{Limitations}
Although we explore the effect of using multi-lingual translations as multiple positives, the experiments are still limited by the number of languages. Study on more low-resource languages could be taken into consideration.
Besides, the XNLI data we used are machine-translated, with possible noises within.
The composition of the training languages and how the training languages can affect the testing on other languages also remains to be explored.
\blue{Additionally, as we study the impact of multiple multi-lingual positives for sentence embedding in this paper, the positives are assigned with equal importance. But there are various fine-grained strategies to weight positives such as ranking deserving exploited.}

\section*{Ethics Statement}
This paper attempts to improve existing sentence embedding approaches. All the data we used are open-sourced and contain no privacy-related ones. Our approaches are based on previously released codebases and checkpoints. We respect all work related to this work and expand ours on their well-established work.
Our work does not introduce ethical biases but aims to make new, positive contributions to the multilingual computational languages community.

\bibliography{custom}
\bibliographystyle{acl_natbib}
\clearpage
\appendix

\section{Appendix}
\label{sec:appendix}
\subsection{Details of Figure~\ref{fig:intro}}
\label{appendix1}
As shown in Figure~\ref{fig:intro}, the similarity relationship in mono-lingual and multi-lingual are different. We calculate the similarity score of sentences by utilizing the released sample code and checkpoint of LaBSE~\footnote{\url{https://huggingface.co/setu4993/LaBSE}}. Some mono-lingual examples we use can be found in Table~\ref{table_example_mono} and multi-lingual examples in Table~\ref{table_example_multi}. 

\blue{For mono-lingual examples, we randomly sample 100 sentences with entailments from the English XNLI dataset~\footnote{\url{https://huggingface.co/datasets/xnli}}. The premise is used as anchor and the corresponding entailment sentence serves as one of the positives. Other positives are generated by ChatGPT~\footnote{\url{https://chat.openai.com/}} and the prompt we used is “Give me several sentences that share similar meaning with the following one: ”. For multilingual examples, we use the same 100 sentences and their corresponding translations from the XNLI dataset and calculate their similarity scores. We report the average similarity score of these sentences in Figure~\ref{fig:intro} (c) and (d).}

\selectlanguage{russian}
\begin{table*}[]
\resizebox{2.05\columnwidth}{!}{
\begin{tabular}{l|l}
\begin{tabular}[c]{@{}l@{}}Anchor\\ Sentence\end{tabular}                     & \begin{tabular}[c]{@{}l@{}}These organizations invest the time and effort to understand their processes and how those processes\\ contribute to or hamper mission accomplishment.\end{tabular} \\ \hline
\multirow{5}{*}{\begin{tabular}[c]{@{}l@{}}Positive\\ Instances\end{tabular}} & These organizations invest lots of time to understand how some processes can contribute to or hamper.                                                                                          \\
& In order to grasp the effects of those processes on the mission, time and effort are spent by these organizations.                                                                             \\
& \begin{tabular}[c]{@{}l@{}}These associations dedicate to comprehending their procedure and assessing how they can either \\ facilitate or impede their mission.\end{tabular}                  \\
& Organizations spent much effort to understand the positive and negative impacts of their processes on the mission.                                                                             \\
& To seize the influence of the processes towards the mission, associations sacrifice much time and effort.                                                                                      \\ \hline
\begin{tabular}[c]{@{}l@{}}Anchor\\ Sentence\end{tabular}                     & \begin{tabular}[c]{@{}l@{}} Thus, with respect to the litigation services Congress has funded, there is no alternative channel for\\ expression of the advocacy Congress seeks to restrict.\end{tabular} \\ \hline
\multirow{5}{*}{\begin{tabular}[c]{@{}l@{}}Positive\\ Instances\end{tabular}} & This is the only channel of expression of the advocacy that Congress seeks to restrict.                                                                                               \\
& Congress intends to curtail advocacy expression through this exclusive channel.                                                                                                       \\
& This channel serves as the exclusive point of restriction for advocacy expression according to Congress.                                                                                               \\
& \begin{tabular}[c]{@{}l@{}} The funded litigation services represent the exclusive channel for the expression of the advocacy\\ Congress seeks to restrict.\end{tabular}                                                                                                  \\                         
& \begin{tabular}[c]{@{}l@{}} With regard to Congress-funded litigation services, there are no alternative means for\\ expressing the advocacy they intend to limit. \end{tabular}

\\ \hline
\begin{tabular}[c]{@{}l@{}}Anchor\\ Sentence\end{tabular}                     & My walkman broke so I'm upset now I just have to turn the stereo up real loud.                                                                                                                 \\ \hline
\multirow{5}{*}{\begin{tabular}[c]{@{}l@{}}Positive\\ Instances\end{tabular}} & I 'm upset that my walkman broke and now I have to turn the stereo up really loud.                                                                                                             \\
& The broken walkman made me feel upset now and I'll turn loud the stereo.                                                                                                                       \\
& I'm feeling upset because my walkman is no longer functional and I'll turn loud the stereo.                                                                                                    \\
& The broken walkman has left me feeling upset, and I have no other choice but to turn up the volume on the stereo.      \\                                                                       
& I'm feeling down because my walkman is broken thus I'll turn loud the stereo.                                                                                                                 
\end{tabular}}
\selectlanguage{english}
\caption{Some examples that we use to calculate the similarity among mono-lingual multiple positive instances. }
\label{table_example_mono}
\end{table*}
\begin{table*}[]
\resizebox{2.05\columnwidth}{!}{
\begin{tabular}{l|l}
\begin{tabular}[c]{@{}l@{}}Anchor\\ Sentence\end{tabular}                     & \begin{tabular}[c]{@{}l@{}}
\selectlanguage{russian}These organizations invest the time and effort to understand their processes and how those processes\\ contribute to or hamper mission accomplishment.\end{tabular}                                       \\ \hline
\multirow{5}{*}{\begin{tabular}[c]{@{}l@{}}Positive\\ Instances\end{tabular}} & \begin{tabular}[c]{@{}l@{}}Diese Organisationen investieren die Zeit und den Aufwand , um ihre Prozesse zu verstehen und \\ wie diese Prozesse einen Beitrag zur Erfüllung der Aufgaben leisten oder behindern.\end{tabular}         \\
& \begin{tabular}[c]{@{}l@{}}Ces organisations investissent le temps et les efforts nécessaires pour comprendre leurs processus \\ et la manière dont ces processus contribuent ou entravent la réalisation des missions.\end{tabular} \\
& \begin{tabular}[c]{@{}l@{}}Estas organizaciones invierten el tiempo y el esfuerzo para comprender sus procesos y cómo esos \\ procesos contribuyen o dificultan el logro de la misión.\end{tabular}                                  \\
& \begin{tabular}[c]{@{}l@{}}
\selectlanguage{russian}Эти организации вкладывают время и усилия для понимания своих процессов и того,
каким образом \\эти процессы способствуют достижению целей миссии или препятствуют их достижению.

\end{tabular}                                                \\
& \begin{CJK*}{UTF8}{gbsn}这些组织投入时间和努力来了解它们的进程以及这些进程如何有助于或妨碍特派团的成就.                           \end{CJK*}                                                                                                                                                                  \\ \hline
\begin{tabular}[c]{@{}l@{}}Anchor\\ Sentence\end{tabular}                     & \begin{tabular}[c]{@{}l@{}} Thus, with respect to the litigation services Congress has funded, there is no alternative channel for\\ expression of the advocacy Congress seeks to restrict.         \end{tabular}                                                                                                                           \\ \hline
\multirow{5}{*}{\begin{tabular}[c]{@{}l@{}}Positive\\ Instances\end{tabular}} & \begin{tabular}[c]{@{}l@{}} So gibt es in Bezug auf den Prozess der Rechtsstreitigkeiten, den der Kongress finanziert hat, \\keinen Alternativen Kanal für den Ausdruck des Advocacy-Kongresses zu beschränken.   \end{tabular}                                                                                                                               \\
 & \begin{tabular}[c]{@{}l@{}}Por lo tanto, con respecto a los servicios judiciales que el congreso ha financiado, no existe ningún canal\\ alternativo para la expresión del Congreso de promoción que pretende restringir.\end{tabular}                                                                           \\
& \begin{tabular}[c]{@{}l@{}}Ainsi, en ce qui concerne le congrès des services contentieux, il n'y a pas de voie alternative\\ pour l' expression du congrès de plaidoyer.\end{tabular}                                                                      \\
& \begin{tabular}[c]{@{}l@{}}

Таким образом , что касается деятельности конгресса по судебным услугам , то не существует\\ какого-либо альтернативного канала для выражения мнений в рамках информационно-пропагандистского \\конгресса .

\end{tabular}                                                                                  \\
& \begin{CJK*}{UTF8}{gbsn}因此, 关于诉讼服务大会提供资金的问题, 没有任何其他渠道可以表达宣传大会试图加以限制的渠道.                                          \end{CJK*}                                                                                                                                                      \\ \hline
\begin{tabular}[c]{@{}l@{}}Anchor\\ Sentence\end{tabular}                     & My walkman broke so I'm upset now I just have to turn the stereo up real loud.                                                                                                                                                       \\ \hline
\multirow{5}{*}{\begin{tabular}[c]{@{}l@{}}Positive\\ Instances\end{tabular}} & Mein Walkman ist kaputt , also bin ich sauer , jetzt muss ich nur noch die Stereoanlage ganz laut drehen .                                                                                                                           \\
& Mon Walkman S' est cassé alors je suis en colère maintenant je dois juste tourner la stéréo très fort                                                                                                                                \\
& Mi Walkman se rompió así que estoy molesto ahora solo tengo que girar el estéreo muy alto.                                                                                                                                           \\
& \begin{tabular}[c]{@{}l@{}}

Мой плеер сломался, так что я расстроен. Мне просто нужно включить стерео погромче.
\end{tabular}                                                                                    \\
& \begin{CJK*}{UTF8}{gbsn}我的随身听坏了所以我现在不高兴了我只能把立体声调大声.                                                 \end{CJK*}                                                                                                                                             
\end{tabular}}
\selectlanguage{english}
\caption{Some examples that we use to calculate the similarity among multi-lingual multiple positive instances. }
\label{table_example_multi}
\end{table*}
\selectlanguage{english}

\subsection{Detailed Results of Main Experiments}
\label{appendix2}
We show detailed results of different tasks of in this section. BUCC's results is shown in \blue{Table~\ref{table_bucc}}. Some detailed results, especially languages involved in our training dataset are shown in \blue{Table~\ref{table_tatoeba}}. STS results can be found in \blue{Table~\ref{table_sts}}. Full results of MTOP Domain Classification are shown in \blue{Table~\ref{table_class}}.

\begin{table}[H]
\centering
\resizebox{\columnwidth}{!}{
\begin{tabular}{lccccc}
\hline
\textbf{Model} & \textbf{fr} & \textbf{ru} & \textbf{zh} & \textbf{de} & \textbf{avg.} \\ \hline\hline
mBERT & - & - & - & - & 56.7 \\
\ \ \ \ + Single  & 85.2 & 83.1 & 80.2 & 88.3 & 84.1 \\
\ \ \ \ + Multiple w/o hard negative & 86.7 & 84.2 & 81.0 & 89.3 & 85.3 \\
\hline
XLM-R & - & - & - & - & 66.0 \\
\ \ \ \ + Single  & 94.2 & 95.1 & 93.1 & 95.2 & 94.5 \\
\ \ \ \ + Multiple w/o hard negative & 94.9 & 96.0 & 95.2 & 96.4 & 95.7 \\
\ \ \ \ + Multiple w/ hard negative & 94.5 & 95.0 & 96.6 & 95.5 & 95.4  \\
\hline
mSimCSE$_{all}$          & -           & -           & -           & -           & 95.2          \\
\ \ \ \ +Single & 94.9 & 96.4 & 96.8 & 96.3 & 96.1\\
\ \ \ \ +Multiple & 95.1 & 96.5 & 96.2 & 96.4 & 96.0\\
\hline
LaBSE          & -           & -           & -           & -           & 93.5          \\ 
\ \ \ \ +Single  & 96.4 & 97.6 & 97.0 & 98.2 & 97.3 \\
\ \ \ \ +Multiple          & 96.9      & 97.8        & 97.6        & 98.1        & 97.6          \\ 
\hline
\end{tabular}}
\caption{\blue{Full results of BUCC task.}}
\label{table_bucc}
\end{table}
\begin{table*}[]
\centering
\resizebox{2.0\columnwidth}{!}{
\begin{tabular}{lcccccccc}
\hline
\textbf{Model} & \textbf{de} & \textbf{fr} & \textbf{es} & \textbf{ru} & \textbf{zh} & \textbf{avg$_{in}$} & \textbf{14 avg.} & \textbf{36 avg.} \\ \hline\hline
mBERT & - & - & - & - & - &- & - &-\\
\ \ \ \ + Single  & 95.8 &89.8 & 88.4 & 86.3 & 87.2 & 89.5 & 70.5 & 64.4\\
\ \ \ \ + Multiple w/o hard negative & 96.2 & 89.8 & 90.1 & 87.9 & 89.1 & 90.6 & 71.6 & 65.1\\
\hline
XLM-R & - & - & - & - & - &- & 57.6 & 53.4\\
\ \ \ \ + Single  & 98.6 &95.0 & 97.6 & 93.6 & 95.6 & 96.1 & 91.3 & 89.6\\
\ \ \ \ + Multiple w/o hard negative & 99.0 & 95.2 & 97.8 & 94.1 & 96.5 & 96.5 & 92.0 & 90.4\\
\ \ \ \ + Multiple w/ hard negative & 98.9 & 95.6 & 97.8 & 94.4 & 96.2 & 96.6 & 93.5 & 91.8\\
\hline
mSimCSE$_{all}$ (Reproduced) & 98.8 & 94.7 & 97.2 & 94.2 & 96.5 & 96.3 & 93.2 & 91.2\\
\ \ \ \ + Single & 98.8 & 95.6 & 98.1 & 93.8 & 96.4 & 96.5 & 93.3 &91.2 \\
\ \ \ \ + Multiple & 99.0 & 95.2 & 98.2 & 94.7 & 96.0 & 96.6 & 93.5 & 91.4\\
\hline
LaBSE (Reproduced) & 99.0 & 96.0 & 96.3 & 95.3 & 96.1 & 96.5  & 95.3 & 95.0 \\
\ \ \ \ + Single &99.2 &96.0 &98.4 & 95.0 & 96.1 & 96.9 & 95.7 & 95.1 \\
\ \ \ \ + Multiple & 99.4 & 96.4 & 98.3 & 95.2 & 96.4 & 97.1 & 96.0 & 95.4         \\ 

\hline
\end{tabular}}
\caption{\blue{Detailed results of Tatoeba dataset}. Since in the paper of LaBSE~\cite{feng2020language} and mSimCSE~\cite{wang-etal-2022-english}, authors did not report the score for each language, we reproduce their scores on Tatoeba dataset through XTREME benchmark to have a better comparison. avg$_{in}$ stands for the languages included in our training dataset.}
\label{table_tatoeba}
\end{table*}
\begin{table*}[]
\centering
\resizebox{2.0\columnwidth}{!}{
\begin{tabular}{lcccccc}
\hline
\textbf{Model} & \textbf{STS17$_{in}$} & \textbf{STS17$_{ex}$} & \textbf{STS17 avg.}  & \textbf{STS22$_{in}$} & \textbf{STS22$_{ex}$} & \textbf{STS22 avg.}\\ \hline\hline

mBERT & - & - & - & - & - &- \\
\ \ \ \ + Single  & 63.7 & 51.5 & 57.0 & 54.3 & 52.2 & 53.4 \\
\ \ \ \ + Multiple & 63.8 & 52.8 & 57.8 & 56.8 & 54.5  & 55.8 \\
\hline
XLM-R & - & - & - & - & - &- \\
\ \ \ \ + Single  & 73.8 & 68.9 & 71.1 & 61.2 & 57.8 & 59.8\\
\ \ \ \ + Multiple w/o hard negative & 74.8 & 71.9 & 73.2 & 63.2 & 58.9 & 61.4 \\
\ \ \ \ + Multiple w/ hard negative & 81.1 & 77.1 & 78.9 & 65.1 & 62.5 & 64.0 \\
\hline
mSimCSE$_{all}$~(Reproduced) & 78.6 & 75.0 & 76.7 & 64.3 &61.5 &63.2  \\
\ \ \ \ + Single & 79.1 & 75.1 & 76.9 & 63.6 &59.8 &62.0  \\
\ \ \ \ + Multiple & 81.0 & 76.3 & 78.5 & 65.4 &62.7 &64.3  \\
\hline
LaBSE & 75.3      & 73.2        &  74.2     &  61.0   &  60.7 & 60.9     \\
\ \ \ \ + Single & 76.1 & 74.6 & 75.3 & 61.0 & 57.8 & 59.7  \\
\ \ \ \ + Multiple & 78.0   & 76.1     & 76.9  &  62.6 &  59.9 &  61.5    \\ 

\hline
\end{tabular}}
\caption{\blue{Detailed results of STS tasks.} We evaluate mSimCSE$_{all}$ through MTEB by ourselves. Task$_{in}$ stands for language pairs, that are inside our training set while Task$_{ex}$ stands for exclusive language pairs. More specifically, in STS17, there are five included pairs and six excluded pairs while in STS22, there are ten and seven, respectively~(excluding \textit{fr-pl}).}
\label{table_sts}
\end{table*}
\begin{table*}[]
\centering
\begin{tabular}{lccccccc}
\hline
\textbf{Model} & \textbf{de} & \textbf{en} & \textbf{es} & \textbf{fr} & \textbf{hi} & \textbf{th} & \textbf{avg.} \\ \hline\hline
mBERT & - & - & - & - & - &- &- \\
\ \ \ \ + Single  & 74.6 & 78.0 & 75.1 & 69.3 & 59.8 & 16.9 & 62.3\\
\ \ \ \ + Multiple  & 74.1 & 78.6 & 76.0 & 69.9 & 60.5 & 16.8 & 62.7 \\
\hline
XLM-R & - & - & - & - & - &- &- \\
\ \ \ \ + Single  & 84.5 & 85.4 & 85.0 & 81.7 & 79.2 & 82.1 & 83.0\\
\ \ \ \ + Multiple w/o hard negative & 86.3 & 86.2 & 86.8 & 82.3 & 83.1 & 82.0 & 84.5 \\
\ \ \ \ + Multiple w/ hard negative & 88.0 & 88.5 & 88.8 & 86.2 & 85.9 & 83.8 & 86.8 \\
\hline
mSimCSE$_{all}$~(Reproduced) & 85.2 & 87.0 & 85.4 &82.6 &	83.0&	81.2&	84.1 \\
\ \ \ \ + Single  & 86.6 &87.7 & 87.4 & 83.8 & 83.3 & 82.1 & 85.2\\
\ \ \ \ + Multiple  & 87.2 & 87.6 & 87.8 & 85.1 & 84.4 & 83.2 & 85.9 \\
\hline
LaBSE & 87.0 & 86.1 & 84.1 & 84.1 & 85.1 & 81.2 & 84.6          \\ 
\ \ \ \ + Single  & 87.2 & 88.0 & 86.2 & 84.5 & 85.7 & 82.0 & 85.6\\
\ \ \ \ + Multiple & 88.0 & 87.9 & 86.7 & 84.6 & 86.5 & 82.7 & 86.1 \\
\hline
\end{tabular}
\caption{\blue{Full results of MTOP domain classification.} We report the accuracy metric on test set. Notice that \textit{hi} and \textit{th} are languages excluded from our training dataset. }
\label{table_class}
\end{table*}

\end{document}